\documentclass{article}

\usepackage[accepted]{icml}

\usepackage[utf8]{inputenc} 
\usepackage[T1]{fontenc}    
\usepackage{hyperref}       
\usepackage{url}            
\usepackage{booktabs}       
\usepackage{amsfonts}       
\usepackage{nicefrac}       
\usepackage{microtype}      
\usepackage{xcolor}         
\usepackage{graphicx}

\usepackage{makecell}

\setcitestyle{square}

\usepackage[colorinlistoftodos]{todonotes}

\begin{document}
\icmltitlerunning{Show and Tell: Enhancing Agent Communication and Learning through Action and Language}

\twocolumn[
\icmltitle{Enhancing Agent Communication and Learning through Action and Language}





\begin{icmlauthorlist}
\icmlauthor{Hugo Caselles-Dupré}{isir}
\icmlauthor{Olivier Sigaud}{isir}
\icmlauthor{Mohamed Chetouani}{isir}

\end{icmlauthorlist}

\icmlaffiliation{isir}{Sorbonne Université, CNRS, Institut des Systèmes Intelligents et de Robotique (ISIR)}


\vskip 0.3in
]

\printAffiliationsAndNotice{\icmlEqualContribution}

%


\section{Introduction}

Reinforcement learning agents are generally considered too slow at learning how to solve problems on their own. One of the standard options to accelerate learning is to consider a teacher-learner pair of agents where the role of the teacher is to drive the learner \cite{sigaud2021towards}. Most work in this domain relies on Learning from Demonstration (LfD) methods such as Inverse Reinforcement Learning and Behavioral Cloning, where the learner copies the demonstrations as they are \cite{argall2009survey}.

Our main point in this paper is to show that one can improve the teacher-learner interaction process by introducing goals in the agents. We thus rely on goal-conditioned agents (GC-agents) which can address a variety of goals \cite{colas2022autotelic}. We show that using GC-agents helps incorporate efficient communication between them, ultimately resulting in pedagogy from the teacher side and pragmatism from the learner side.

In more details, when goals are introduced, the teacher can convey goals through demonstrations and instructions.
When given a demonstration, the learner needs to figure out the goal of that demonstration, enabling it to generalize the communicated knowledge of the demonstration without simply copying it \cite{ho2016showing,casellespragmatically}. 
Similarly, the integration of instructions and language as additional communication modalities becomes particularly meaningful when combined with the concept of goals. The teacher can use instructions to convey not only what actions to take but also the underlying subgoals to achieve. This allows for a more comprehensive and nuanced form of communication, where the teacher can provide guidance on how to approach different goals, handle various scenarios, and adapt strategies accordingly \cite{caselles2022overcoming}.


We thus introduce a novel family of GC-agents that can act both as teachers and learners, leveraging action with demonstrations and language with instructions as two modalities for communication. We demonstrate how these agents can utilize their goal space to enhance communication, learning, and efficiency. Specifically, we explore the implementation of pedagogy \cite{ho2019communication} and pragmatism \cite{shafto2014rational}, two communication mechanisms essential for effective human communication and goal achievement \cite{gweon2021inferential}. By incorporating these qualities into our agents' communication strategies, we enhance their ability to teach and learn effectively. Furthermore, we investigate the impact of mixing communication modalities (action and language) on learning performance, demonstrating the benefits of a multi-modal approach.

\section{Methods}

\begin{figure}[ht!]
    \centering
    \includegraphics[scale=0.25]{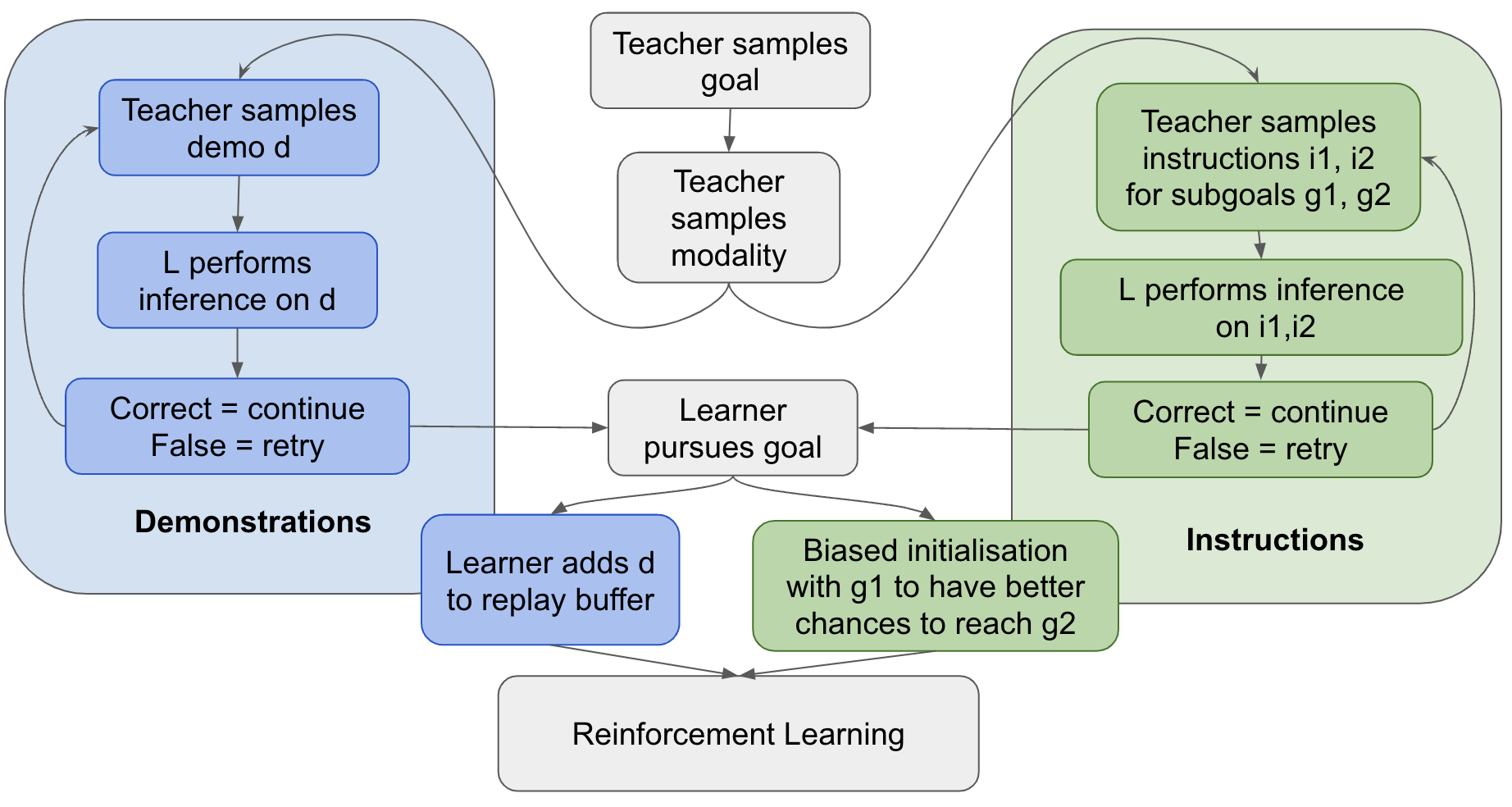}
    \caption{Learner training loop summary.}
    \label{fig:methods}
\end{figure}

\begin{figure*}[ht!]
    \centering
    \includegraphics[scale=0.28]{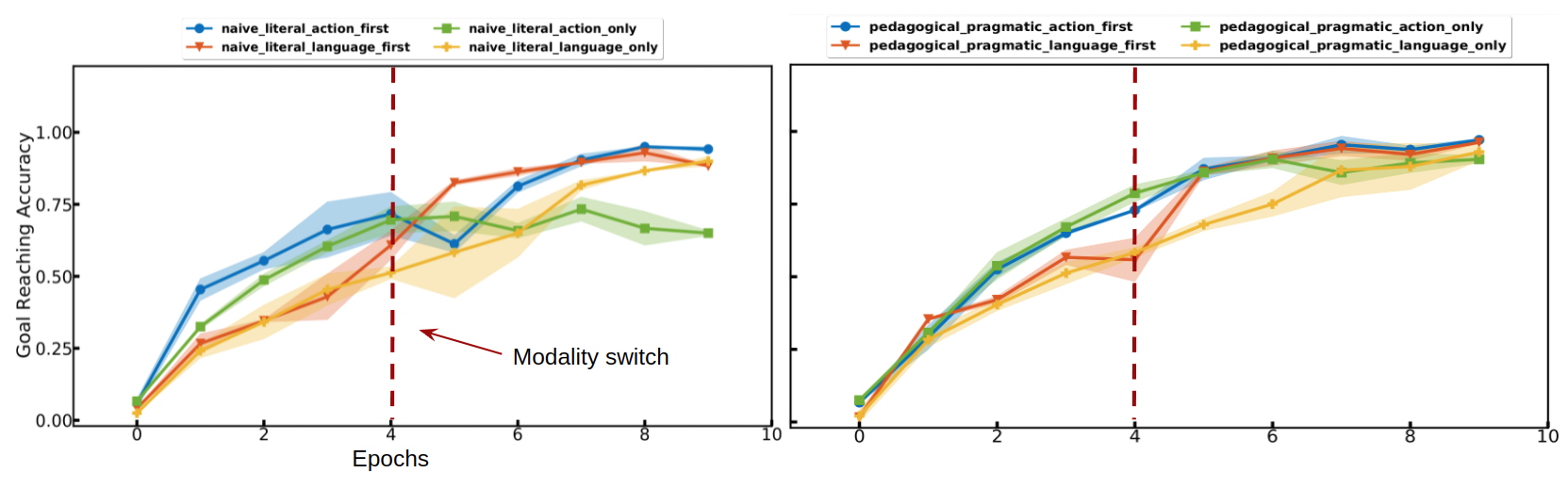}
    \caption{Left: Naive Teachers with Literal Learners. Right: Pedagogical Teachers with Pragmatic Learners.}
    \label{fig:results_nl_pp}
\end{figure*}

In our study, we compare two different methodologies for teaching and learning in the context of a Fetch robot learning to stack three cubes in various configurations, including pyramids and stacks of three cubes. Our setup is derived from the Fetch Block Stacking environment \cite{caselles2022overcoming, casellespragmatically}, and thus uses the same metrics as well as the same environment. 

The first approach involves a naive teacher, which provides demonstrations without any specific goal prediction, and a literal learner, which learns directly from these demonstrations. The second methodology introduces a pedagogical teacher that learns to predict the goals of its demonstrations, enabling more targeted instruction, and a pragmatic learner that learns to predict the goals of its own behavior, enhancing its preparation for communication. 

These methodologies are explored using two modalities: action with demonstrations, where the teacher physically demonstrates the desired stacking behavior, and language with instructions, where the teacher provides verbal instructions for the learner to follow. 

The communication process between the teaching and learning agents operates as follows: the teacher agent delivers a teaching signal in the form of either a demonstration or an instruction. The learner agent, employing Bayesian Goal Inference, attempts to predict the underlying goal associated with the teaching signal. It iteratively retries until successfully inferring the goal, after which it utilizes the teaching signal to its advantage. For demonstrations, the learner adds them to its replay buffer, enabling it to learn from the demonstrated behavior. In the case of instructions, the learner receives a curriculum of subgoals, aiding its understanding and progress. Subsequently, the learner agent actively pursues the predicted goal and updates its policy using reinforcement learning techniques. This learning protocol is visually described in Fig.\ref{fig:methods}

\section{Results and Conclusion}

\paragraph{Results.} Our results show the best choices of modalities (action or language) when considering teaching and learning methodologies (pedagogy and pragmatism) and teaching timing (when is the teaching signal delivered). 


\paragraph{How does teaching and learning methodologies inspired from Developmental Psychology (pedagogy and pragmatism) impact the efficiency of GC-agents?}

As seen in our results in Fig.\ref{fig:results_nl_pp} and Tab. \ref{tab:auc}, the addition of pedagogy and pragmatism increases learning efficiency in AI agents for all tested modalities. Whether the agents learns from demonstrations only, instructions only, with demonstrations first and then instruction or vice-versa, all agents benefit from teaching and learning methodologies present in human teaching and learning. 

Adding such teaching and learning methodologies provides an additional 2 points in Area Under the Curve on average. It provides the most gains when the agent only learns with demonstrations. Those results indicate that efficient human teaching and learning methodologies transfer to GC-agents.

\begin{table}[ht]
  \caption{Quantitative results on comparing modalities and combinations of teachers/learners.}
  \label{tab:auc}
  \centering
  \begin{tabular}{lll}

    & \multicolumn{2}{c}{Training efficiency (AuC)}  \\
    \cmidrule(r){2-3}
    \thead{Modality}     & \thead{Naive Teacher + \\ Literal Learner}     & \thead{Pedagogical + \\ Pragmatic} \\
    \midrule
    \thead{Demonstrations only}     & \thead{28,8} & \thead{\textbf{35,7}} \\
    \thead{Instructions only}     & \thead{28,3} & \thead{\textbf{30,6}} \\
    \thead{Demonstrations then \\ Instructions} &  \thead{34,3} &   \thead{\textbf{36,2}} \\
    \thead{Instructions then \\ Demonstrations} & \thead{31,2} & \thead{\textbf{34,2}}     \\

    \bottomrule
  \end{tabular}
\end{table}

\paragraph{Should GC-agents teach by showing (demonstrations) or telling (instructions)?} 

Our experiments then reveal that mixing the nature of teaching signals is beneficial to GC-agents. Teaching with demonstrations and instruction, no matter the order in which they are given, is superior to learning with demonstrations only or instructions only.  When mixing the nature of teaching signals, the question of order and timing is raised. Our experiments reveal that an artificial teacher should choose a demonstration rather than an instruction at the beginning 
of teaching, and use instructions rather than demonstrations later on. We land on the same conclusions as in Developmental Psychology research: demonstrations, which provide low-level information in the form of examples work well early on, which instructions and more generally language allows to generalize and provide information about general concept rather than particular examples \cite{sumers2020show}.

\paragraph{Conclusion.} In conclusion, our work shows the role of goals in the communication and learning processes of goal-conditioned agents. By incorporating pedagogy and pragmatism as methodologies, we enable effective teaching and learning dynamics. Furthermore, our experiments reveal the benefits of using and mixing communication modalities (action and language) in GC-agents.

\section*{Acknowledgement}

This work was performed using HPC resources from GENCI-IDRIS (Grant 2022-A0131013011).


\bibliography{bibli}
\bibliographystyle{icml}

\end{document}